\documentstyle{article}
\title{The Very Idea of Dynamic Semantics: An Overview from the Underground}
\author{David Israel \\ Artificial Intelligence Center \\ SRI}
\date{}

\newcommand{\EE}{{\bf $\exists$}E}
\newcommand{\UI}{{\bf $\forall$}I}
\newcommand{\EI}{{\bf $\exists$}Inst}
\newcommand{\UG}{{\bf $\forall$}G}

\begin{document}

\maketitle

\section{Introduction}

Imagine someone complained as follows: ``Our standard accounts of the
semantics of Natural Languages are sadly inadequate, and a crucial
source of their inadequacy is that they do not take account of or
reflect the dynamics of language; they are simply too static.''  One
might well agree, yet wonder more exactly what the complainant was
talking about.  Thinking of the work of Wittgenstein, Austin, Searle
and others, one might respond, ``Oh, you mean that inadequate
attention has been paid to uses of language other than the
statement-making use; that standard semantic accounts have ignored all
the kinds of things we do with language.''  But, no, that's not the
complaint they have in mind; the plaintiff is perfectly happy
accepting the centrality of the statement-making use of language and
the correlative centrality of indicative sentences.  One might then
reflect on the fact that the roots of our standard accounts are to be
found in the work of Frege, Russell, and others on the logic of
mathematics and that there are no dynamics within mathematical
structures.  The domain of mathematics is a domain of unchanging
entities and unchangeable relations, but we most often use natural
languages to describe and report on things that change and act on one
another.  But, no, you are told, this is not the source of the
dissatisfaction, either.

Finally you decide to ask for some examples, some phenomena that
motivate the complaint---phenomena no adequate account of which can be
had without moving to a more dynamic, less static framework.  Wouldn't
you be surprised when given the following list of examples?

\begin{enumerate}
\item If Hans owns a donkey, he pets it.
\item Every semanticist who owns a donkey pets it.
\item A donkey came to my office this morning.  It had a theory about
people anaphora.
\end{enumerate}

Perhaps the third example does suggest something, something about the
importance of sequences of sentences, but otherwise, you would be at a
loss---unless, of course, you knew a lot about work in natural
language semantics over the last decade or so.  So, what is the
problem with these sentences and for whom is it a problem?  Whose oxen
do these sentences gore?

\subsection{The Phenomena}

First let us make clear the intended, but troublesome readings of our
sentences

\begin{enumerate}
\item If Hans owns {\em a donkey}$_{i}$, he pets {\em it}$_{i}$.
\item Every semanticist who owns {\em a donkey}$_{i}$ pets {\em it}$_{i}$.
\item {\em A donkey} came to my office this morning.  {\em It}$_{i}$
had a theory about people anaphora.
\end{enumerate}

On the readings intended, the occurrences of `it' in these sentences
are [meant to be] semantically dependent on the corresponding
(co-indexed) indefinites.  The problem is that it's not clear that, or
how, this dependence is mediated by the syntactic structures.
Consider, that is, the following diagnosis:

\begin{quote}
The phenomena all involve {\em anaphoric pronouns} {\em outside the
scope} of their [indefinite] NP antecedents.
\end{quote}

\noindent One might wonder whether all cases of
semantic dependence of pronouns on NP's of various kinds are to be
classified as anaphoric; in any event, the theory-ladenness of this
description is carried by its use of a notion of scope.  Where is the
operator whose scope is in question, and what is that scope?

Part of the intended formal model is clear enough; anaphoric pronouns
are to be modeled by variables and their quantified antecedents are to
be modeled by some form of variable-binding quantificational operator.
There are various possibilities.  We won't go into any in any detail;
we simply want to note that it is not at all obvious that there are
any {\em variables} in English.  There are, of course, pronouns,
though not many: nobody seems to treat English as an n-variable
fragment (for relatively small n) of some standard quantificational
language.  If there are no variables, then there are no
variable-binding operators, either.  (Or they would all have only
vacuous occurrences.)  We can grant that pronouns have uses on which
they are semantically dependent on other noun phrases in ways that are
similar to, and for certain purposes can be adequately modeled by.
the relation between a variable $x_{i}$ and a variable-binding
quantificational operator $Qx_{i}$ that binds it.  But this means no
more than than what it says.  In particular, one should be careful
about identifying elements/aspects of a model, even a very good one,
with elements/aspects of the phenomena modeled.  So one should be
careful about claiming that there are variables and
variable-binding operators in English.

In the foregoing, we have been ignoring the existence of real
honest-to-goodness variables in informal mathematics, physics, etc.
Consider, ``Let {\em A} and {\em B} be sets; then $A = B$ if and only
if, for each $x$, $x$ is a member of $A$ if and only if $x$ is a
member of $B$.''  By my lights, this is English all right, and $A, B,
x$ are variables. But it is well to remember that the use of variables
in informal mathematics is ambiguous and that the interpretation of
variables, context dependent. If in the course of a discussion of
integers, the formula $x + 7 = -1$ appears, we are almost surely meant
to understand the variable as existentially quantified.  If we see
instead $x + 7 = 7 + x$, we will take the quantification to be
universal.  Finally, if we see $x + y =z$ in that same discussion, we
won't know how to take it---except that we won't take all three as
universally quantified---without further hints from the context.

Even assuming that there are variables and variable-binding operators,
it is not clear that there is a unique, well-defined notion of the
{\em scope} of these operators.  One way of understanding the
phenomenon of donkey anaphora is precisely this: One can opt for a
precise notion of scope, say as determined by something like
c-command, but then also note that a quantifier phrase can have
pronouns or anaphoric definite descriptions coreferential with or
dependent on it that are arbitrarily far from it, not even within the
same sentence, and hence that being within the scope of such a phrase
is not a necessary condition for being anaphorically dependent on it.
Donkey anaphors or discourse anaphors are not bound by those phrases,
though semantically dependent on them.  This motivates accounts that
conceive the semantics of anaphoric but unbound pronouns and anaphoric
definite descriptions as on the model of demonstratives, but
demonstratives with the special feature that the entity to which they
are directed must be made salient by linguistic means.  Or one can do
what
\cite{GroenendijkStokhof91}, \cite{GroenendijkStokhof90},
\cite{PaginWesterstahl93}, and  \cite{Kamp81} do and alter the syntax
(and semantics) of certain variable-binding quantifiers, by altering
the notion of their scope or their selectivity, degree, etc.  Or one
can provide two notions of scope: an analogue of the standard
syntactic one, the smallest wff.  following the operator---this is
akin to the notion of c-command---and a semantic one, {\em semantic
scope}, capturing the observed facts about dependency, but quite
undetermined by syntactic structure.

\section{Historical Background}

Formal languages were first devised precisely to overcome what were
felt to be the inadequacies, for certain delimited purposes, of
informal, natural languages.\footnote{The notion of a formal language
as used here is not a precise technical notion.  We have in mind
formalisms devised by particular humans at particular moments of
history, for some practical or theoretical purpose, and amenable---and
intended to be amenable---to precise mathematical characterization, at
least with respect to certain syntactic properties.} In particular,
the original purpose was to investigate the logical foundations of
mathematics and more generally, to analyze systematically logically
valid inferences and to delimit the logically valid from the invalid.
Thus, Peano, in the Preface to ``The Principles of arithmetic,
presented by a new method,'' says:

\begin{quote}

Questions that pertain to the foundations of mathematics still lack a
satisfactory solution.  The difficulty has its main source in the
ambiguity of language...I have denoted by signs all ideas that occur
in the principles of arithmetic, so that every proposition is stated
only by means of these signs...With these notations, every proposition
assumes the form and the precision that equations have in algebra;
from the propositions thus written other propositions are deduced, and
in fact by procedures that are similar to those used in solving
equations.  This is the main point of the whole paper.
(\cite{vanHeijenoort67}, p.85)

\end{quote}

Frege, in the Preface to his ``Begriffschrift,'' notes that he had set
out to determine how much of arithmetic could be established by logic
alone:

\begin{quote}
To prevent anything intuitive from penetrating here unnoticed, I had
to bend every effort to keep the chain of inferences free of gaps. In
attempting to comply with this requirement in the strictest possible
way, I found the inadequacy of language to be an obstacle...This
deficiency led me to the idea of the present ideography. Its first
purpose, therefore, is to provide us with the most reliable test of
the validity of a chain of inferences and to point out out every
presupposition that tries to sneak in
unnoticed...(\cite{vanHeijenoort67}, p. 5f)

\end{quote}

It is quite clear that one central purpose of these new formal
languages was to enable one to state effective tests of the property
of being a proof and the relation of a sentence being correctly
inferred from others.  Correlatively, a central inadequacy of ordinary
language was that one could not do this for inferences stated in such
language---the language of life (``Sprache des Lebens'').  Frege goes
on to explain the relation between his formalism and ordinary
language.

\begin{quote}

I believe that I can best make the relation of my ideography to
ordinary language clear if I compare it to that which the microscope
has to the eye.  Because of the range of its possible uses and the
versatility with which it can adapt to the most diverse circumstances,
the eye is far superior to the microscope.  Considered as an optical
instrument, to be sure, it exhibits many imperfections, which
ordinarily remain unnoticed only on account of its intimate connection
to our mental life.  But, as soon as scientific goals demand great
sharpness of resolution, the eye proves to be insufficient.  The
microscope, on the other hand, is perfectly suited to precisely such
goals, but that is just why it is useless for all others.

This ideography, likewise, is a device invented for certain scientific
purposes, and one must not condemn it because it is not suited to
others.  (\cite{vanHeijenoort67}, p. 6)

\end{quote}

\noindent In the first sections of work, Frege returns to this theme again, and
in a way that is especially relevant to the concerns of this essay:

\begin{quote}

In ordinary language, the place of the subject in the sequence of
words has the significance of a {\em distinguished} place, where we
put that to which we wish especially to direct the attention of the
listener.  This, may, for example, have the purpose of pointing out a
certain relation of the given judgment to others and thereby making it
easier for the listener to grasp the entire context.  Now all those
peculiarities of ordinary language that result only from the
interaction of speaker and listener---as when, for example the speaker
takes the expectations of the listener into account and seeks to put
them on the right track even before the complete sentence is
enunciated---have nothing that answers to them in my formula language,
since in a judgment I consider only that which influences its {\em
possible consequences}.  Everything necessary for a correct inference
is expressed in full, but what is not necessary is generally not
indicated; {\em nothing in left to guesswork}.
(\cite{vanHeijenoort67}, p. 12.  All emphases in the original.)

\end{quote}

Logical formalisms, on this view, are analytical tools (``scientific
devices'') created to help scientists study a certain range of
phenomena.  They are more than this, though: they are analytical tools
by way of being models of the phenomena to be analyzed.  In the case
at hand the phenomena included aspects of the expression of thoughts
about mathematical entities and, most centrally, certain features of
proofs in mathematics.\footnote{It is clear enough that the languages
of informal mathematics in which mathematics is actually done and in
which results are presented and explained are annexes to ordinary
language.} Of course, neither Frege nor Peano thought that proofs in
their formal systems were adequate models of any other feature of
proofs in mathematics beyond that of exhibiting in a systematic and
effectively determinable way the relation of logical consequence
between axioms/postulates and theorems.

\section{Programming Languages as Analogues, not Metaphors}

We noted in \S 3, that formal languages were objects to be studied, and
to be used in the study of natural phenomena, but that they were not
devised to be used directly to communicate.  There are, though, formal
languages that have been devised precisely for the purposes of
communication: programming languages. Of course, these are used to
communicate with computers primarily and with other humans only
secondarily, and for many purposes the communication is thought of as
only-one way.  Despite this, and despite their view of the
computational metaphor, Groenendijk and Stokhof point to work on the
logic of programs, in particular to Dynamic Logic, as a source of
insight and technical detail for their own contributions.

Dynamic Logic (\cite{Pratt76,Harel79}) is a formal system for
reasoning about and proving properties of programs from languages
designed to be executed by state machines.  Such programs can, in
general, be thought of as associated with transformations of states;
starting from an initial (input) state, the execution of a program
will cause the machine to go through a sequence of states, and perhaps
to halt in a final (output) state.  Dynamic logic abstracts from the
intermediate states and associates with a program a binary relation
between input and output states.  As for the states themselves, in
Propositional Dynamic Logic these are completely abstract atoms, as
they are in the most abstract version of Kripke-style semantics for
modal logics.  Indeed, PDL is a multi-modal logic, with each atomic
program determining an associated normal accessibility relation.  In
First-Order Dynamic Logic, states are modeled by first-order
assignments or valuations and programs determine binary relations
between such valuations.  This idea reappears in Dynamic Predicate
Logic.

In Dynamic Logic, these binary relations model actual state
transformations; in particular, those transformations associated with
the central programming construct of the class of programming
languages under study.  That construct is the simple {\em assignment}.
Where $x$ is a program variable or {\em identifier} and $e$ is
e.g., a numerical expression, an assignment takes the following form;

\begin{center}
\begin{tt}
x := e
\end{tt}
\end{center}

Identifiers refer to abstract {\em locations} (cells, registers);
these are the components of the store or memory of a state machine.
Assignments change state (store, memory).  After executing the
assignment command above, the state of the machine will be just as it
was before except possibly with respect to the value stored in the
cell named or referred to by $x$; the {\em contents} of $x$ will be
the value of the term $e$.\footnote{It should be obvious how this
understanding of assignments relates to the treatment of
quantification in first-order logic; the most direction connection,
however, is in terms of so-called {\em random assignments}, a
nondeterministic assignment in which an arbitrary member of a certain
data type is assigned to a location.  See below.}

Notice that variables can also occur on the right hand side:

\begin{center}
\begin{tt}

x := x + 1

\end{tt}
\end{center}

The occurrence of $x$ on the right is unlike that of $x$ on the left.
The term $x + 1$ does not refer to the successor, in some fixed
ordering, of the location referred to by $x$.  Rather the term on the
right is a numerical expression, referring to the number which is the
successor of the number that is the content (value) of location $x$
prior to the execution of the assignment.\footnote{In this respect, the
occurrence on the right is akin to that of a free variable being used
to model some referential relation between an utterance of a sentence
with a context-dependent element, like a demonstrative, and the object
referred to by that element in that utterance.}

Most real state-based or {\em imperative} programming languages
provide constructs, called {\em declarations}, for allocating and
initializing locations, often restricted by sorts or data types, as
follows:

\begin{center}
\begin{tt}
begin {\bf integer} x; \\
\hspace*{0.5in}   x := 7; \\
\hspace*{0.6in}   x := x$^{2}$; \\
\hspace*{0.6in}   print(x) \\
\hspace*{-0.7in} end
\end{tt}
\end{center}

Here we also have an instance of a {\em block}, in this case a simple
instance in which a new variable is declared with an initial value and
then a sequence of simple commands follows which have that variable as
a (so-called) formal parameter.  These commands constitute the {\em
scope} of the variable.\footnote{One may compare explicit block
structure with Groenendijk and Stokhof's $\Diamond$ operator.} Here
the scope is easily determinable from the program text.  Things are
not always so simple.  Some programming languages have dynamic scoping
rules governing parameters in procedures, so that the arguments to a
procedure are only determined at run-time, when the procedure is
invoked.  More to the point, a given syntactic characterization of a
language, e.g., via an inductive definition of the set of well-formed
programs, does not determine the nature of the scope mechanisms.  As
we shall see below, this determination can be provided in various
ways; roughly such a determination is provided by specifying how
machines to execute the programs are supposed to operate.\footnote{The
possibility of dynamic vs. lexical or static scoping also arises in
purely functional languages as well.  See \cite{EijckFrancez93} for a
use of this difference, in an imperative programming language
paradigm, in modeling the availability of alternative interpretations
of ellipsed verb phrases.} Still, here let us assume that the scope of
a variable is always determinable {\em statically}, from the text of
the program.  The same is not true of the {\em extent} or {\em
lifetime} of a variable or location.

We have said that declarations allocate and initialize a location; the
extent of a variable (or of a variable declaration) is its
computational duration---the length of (notional) time during which
the allocation associated with the declaration is in force.  In some
programming languages, the extent of a variable ends when the control
exits the block in which the variable was declared; it lasts through
temporary jumps out of the block, however.  In other languages,
explicit constructs ending extents (releasing or deallocating the
location) are provided.  In some though, extents are indefinite and
cannot in any way be determined statically from the program text.  In
such cases, the implementation of the language will provide {\em
garbage collection} facilities for retrieving those locations that
have become inaccessible for reuse.

Indeed what we just said about extents, or what we might have said
about dynamically scoped variables, can also be said about {\em
semantic scope}.  Consider the scope of the existential quantifier,
according to Groenendijk and Stokhof.  It is not determinable
statically, from the text or discourse---at least not until we have
reached the end of the text or discourse, and this need not be
determinable from any point in the text short of the end.

Though Groenendijk and Stokhof reject the computational metaphor, van
Eijck and De Vries seem to accept it and accept, in particular, an
interpretation of PDL according to which assignments, in the sense of
valuations, model aspects of the cognitive states of listeners and
natural language sentences are, like programs, associated with
transformations of cognitive states.  Natural language sentences are
akin to programs, executed, that is uttered, so as to bring about
changes in the cognitive states of one's audience.\footnote{This view
can be seen as a high-tech version of Grice's perspective on language
and meaning.}

Perhaps a closer analogy to program variables or locations is to be
found in Kamp's {\em discourse markers} or Heim's {\em indices}.  One
can think of the discourse representation construction algorithm as a
program that, typically at least, begins by declaring a number of
(sorted) variables, with (perhaps) determinate scopes but
indeterminate extents. The assignments involved in the
initializations, however, are sometimes, e.g., when associated with an
indefinites {\em an N}, not simple, but rather {\em random
assignments}, of the form:

\begin{center}
\begin{tt}
begin  x$_{{\small N}}$; \\
\hspace*{0.75in}	x := ?;

\end{tt}
\end{center}

\noindent whose meaning command is to assign a random value of sort {\bf N} to
$x$.

Notice that we have not supplied an {\tt end} for this block,
precisely because in general this can not be determined unless we are
given the text or discourse as a whole.  Of course, this means that
not only is the extent of $x$ indeterminate, so too is its scope.  In
our analytic practice, in which we are presenting theories or models
of the use of natural language, we can stipulate that even though our
object of study is, e.g., the understanding of discourses or texts as
we (as subjects) encounter them, we (as scientists) have access to the
completed discourse or text.  Still, we must realize that one
discourse is or can followed by another, after interruptions no doubt,
where the later is about and is intended to be recognized to be about
the very same objects as the earlier. Does what we mark as the end of
a discourse or text mark the end of the extent (or scope) of a
variable?  Or do we rather rely on something akin to a garbage
collection mechanism for retrieving free cognitive cells?  This
question is connected, of course, to the question of the accessibility
of sub-DRS's within the single DRS that results from processing a
given text or discourse.  But in the present context, these odd
questions are really about the human cognitive architecture, in
particular about interactions between short-term, on-line processing
of linguistic information and long-term memory structures.  Even more
particularly, they are questions about the management of relatively
persistent notions of individuals about whom we gain information in
many different ways and over extended and interrupted periods of time.

\subsection{Extending the Analogy}

Of course it can not be surprising that any theory of natural language
use that involves reference to the processes of understanding will
involve reference to our cognitive architecture.  The same is true for
all fully adequate accounts of programming languages.

Programming languages can be thought of, at least partially, on the
model of more familiar formal languages for mathematics.  This is
especially true for so-called functional languages; one can think of
the meaning of a program (or term) in such a language as denoting some
entity in a mathematical structure---as opposed to an algorithm or set
of processes or sequence of actions---and of the denotations of
complex terms in general as some function of the denotations of their
constituents.  This is the project of {\em denotational semantics} of
programming languages.  Even for functional languages, however, the
question of the computational content of such a semantic account
arises.  For, as we noted above, programming languages are languages
meant to be used, by us, to control the behavior of other
entities---in this case, of computers.  Thus, the equations of a
denotational semantic account are often themselves conceived of as
programs in a functional programming language whose modes of execution
must then be specified in some more operational way.  Or these
equations may be given a direction in a term rewriting system and this
latter operationalized.

Quite generally, then the design of a programming language is at the
same time the design of an abstract machine or family of machines to
execute programs in that language.  Of course, the level of
abstraction necessary and useful for an implementor of the language
(on a real, commercially available) machine is quite different from
that useful to a programmer.  More directly, perhaps, the design of a
programming language involves the specification of the behavior we
intend to induce by the execution of programs from that language.
This specification, whether by way of an abstract machine, by way of
defining the transitions between abstractly characterized machine
configurations, or by way of a definition of the relation between
configurations and results (values) constitutes an {\em operational
semantics} for the programming language.

We have just urged that a fully adequate semantics for a programming
language will involve both a denotational and an operational
semantics; but, of course, this is not enough.  Surely we want to
relate the two accounts in some systematic and illuminating way.  One
way is as follows.  We can assume that the denotational account will
provide a notion of denotational equivalence for programs; so where
$\tau_{1}, \tau_{2}$ are programs, and $[\![ \cdot ]\!]$ is the
semantic function specified by the account, we will have cases in
which $[\![\tau_{1}]\!] = [\![\tau_{2}]\!]$.  As hinted above, things
are a bit more complicated on the operational side.  To simplify,
though, we can say that two programs are operationally equivalent when
they can be substituted, one for the other, in a larger context, a
larger encompassing program, without altering the behavior of that
larger program.  So we assume we can delimit a set of program contexts
$C[\,]$ and propose the following (schematic) definition:

\begin{itemize}
\item {\bf Operational Equivalence} Two {\em expressions} $\tau_{1}, \tau_{2}$,
are operationally
equivalent iff they induce the same {\em behavior} in all program
contexts---iff for any program context, $C[\,]$,
\begin{center}
$C[\tau_{1}] =_{oper} C[\tau_{2}]$
\end{center}
\end{itemize}

Then the desired fits between operational and denotational semantics
are called {\em correctness} and {\em full abstractness}:

\begin{itemize}

\item {\bf Correctness} $[\![\tau_{1}]\!] = [\![\tau_{2}]\!]
\Rightarrow C[\tau_{1}] =_{oper} C[\tau_{2}]$
\item {\bf Full abstractness}  $C[\tau_{1}] =_{oper}
C[\tau_{2}] \Rightarrow [\![\tau_{1}]\!] =  [\![\tau_{2}]\!]$

\end{itemize}

The requirement, in particular, of full abstractness is rather like
that of compositionality\footnote{For which, see below
Section~\ref{sec-conc}.}: it is a methodological ideal---a good thing
if you can get it, but it can be quite hard to attain and the price
can be too high.  There are relatively easy cases, though.  Indeed, as
we move away from functional programming languages, we may find that
the gap between natural denotational accounts and illuminating
operational accounts narrows; in particular, it may be that the
underlying mathematical space in which the denotations of programs is
to be found is best conceived of as consisting of fairly abstractly
conceived configurations.\footnote{The standard approach to proving
full abstractness for imperative languages within the denotational
paradigm consists in defining a typed $\lambda$-calculus (with fixed
points) as a metalanguage, e.g., with explicit functions over stores
or states and providing semantics for that language in terms of a
particular family of mathematical structures called {\em domains}.}
One such example is precisely that of the standard
state-transformation, `dynamic' semantics for sequential
while-programs, which is provably fully abstract with respect to the
partial correctness behavior of such programs, as axiomatized for
instance, by Propositional Dynamic Logic or by way of Hoare triples.
A partial correctness assertion about a program states that if the
program is executed in a state that meets a certain condition
(involving the values assigned to locations/variables), then a certain
other condition will be true if and when the program terminates.

Van Eijck and de Vries use Hoare logic to reason about the `programs'
expressible in Dynamic Predicate Logic, which they, unlike Groenendijk
and Stokhof, understand in a straightforwardly computational way.  In
a sense, their soundness and completeness results\footnote{for both
partial and total correctness assertions} establish correctness and
full abstraction for PDL with its the state-transformation semantics.

\subsubsection{An Abstract Machine for DRT?}

In our discussion of correctness and full abstractness, we noted that
one must specify some notion of behavior and of behavioral equivalence
and that one could do this in many different ways and at many
different levels of abstraction.  In the case mentioned above, of
partial correctness behavior, we are interested only in the
input-output behavior of programs, with respect only to highly
abstract properties of undecomposed states.  To get a feel for what
might be involved in a more ful-bodied conception of behavior, let us
turn briefly to a consideration of Kamp's conception, which, though
certainly not straightforwardly computational, is `mentalistic' and
oriented toward issues of cognitive processes.

One can think of the discourse representation construction algorithm
as specifying the reader or parser for an interpreter for an abstract
machine whose input is a stream of sentences and whose output (when it
is initialized with a starting DRS) is a stream of DRS's or a single
over-all DRS for a given discourse or text.  In what follows, we shall
presume that for a given text or discourse, {\em however this is
delimited}, a single DRS is produced.  So the reader produces a new
DRS, $DRS^{\prime}$ given a sentence and the DRS $DRS$ produced so
far.  The additions to $DRS$ are determined by the parsing of the
sentence into constituents, the commands that each type of constituent
triggers, together with the input $DRS$.

We might characterize the set of configurations $\Gamma$ for a DRT
machine as follows:

\begin{center}
$\Gamma = \langle Cst^{*}, Com^{*} \rangle \times DRS \times Val^{*}
\times DRS^{\prime}$
\end{center}

\noindent where $Cst$ is the set of types of constituents, $Com$ is
the set of commands, $DRS$, the set of discourse structures $DRS$ =

\begin{center}
\[ DRS = \langle DM, Con \rangle \]
\end{center}

\noindent  and  $Val = DM + Con$.   If there are no more constituent-command
pairs on the stack, the system terminates, with $DRS^{\prime}$ as the
final result.  The stack of values is for holding intermediate results
of the interpretation of subsentential constituents.  The questions
raised before about the extent of discourse markers would be answered
by a detailed specification of an interpreter for this abstract
machine.

Thus imagine that one thought of giving an operational semantics for
{\em English} by way of specifying the discourse representation
construction algorithm as an interpreter for a state machine (our
minds) whose configurations were as sketched above.  One would want to
specify some appropriate notion of behavior and of behavioral
equivalence.  Here's a first very simple idea.  Where texts are finite
sequences of sentences:

\begin{itemize}
\item {\bf Operational Equivalence} Two {\em sentences} $s_{1},
s_{2}$, are operationally equivalent iff they induce the same {\em
final DRS} in all texts---iff for any text, $C[\,]$,
\begin{center}
$C[s_{1}] =_{oper} C[s_{2}]$
\end{center}
\end{itemize}

\noindent where $C[\,]$ is the DRS that results from executing the DRS
construction algorithm, initialized with the empty DRS, on the text
$C$.

Of course, one may have to specify much else to yield any determinate
analysis in this regard: in particular the perceptual capacities of
the agent and the nonlinguistic knowledge base and knowledge base
routines. e.g., access and inference routines, exploited by the
language understanding algorithm.  Or perhaps not: perhaps there is
enough of an autonomous language-understanding module to justify talk
of the purely linguistically accessible content of a text---what John
Perry and I call the purely reflexive semantic content of the text.
Waving all that aside, do we have any reason for thinking that there
are many operationally equivalent sentence pairs?  Perhaps the
sentence is the wrong `unit'; perhaps only larger (or smaller)
stretches of text will do.  What category or categories of linguistic
expression correspond to programs---to items that are directly
executable and whose execution leads to observable behavior,
observable transformations of configurations?

\section{Proofs as Another Source of Analogies}

Let us return to a simple instance of the most straightforward example
of so-called discourse anaphora:

\begin{quote}
A man walked in.  He sat down.

\end{quote}

Heim, Kamp, Van Eijck and de Vries all suggest something like the
following processing story:

\begin{enumerate}

\item The processor is to focus on an {\em arbitrary} man, call him $a$,
\item such that $a$, {\em the man chosen}, walked in;
\item then the processor is to add to the information about that man that
he sat down.
\end{enumerate}

The process in question is akin to that of opening up a file for a
man, labeling that file $a$ and incrementally entering new information
about whomever it is that is called $a$.  We have urged above that
that process is analogous in certain ways to declaring, allocating and
initializing a location named by a variable and then invoking
procedures involving that variable.  It is also analogous to certain
steps in the construction of proofs---but to what steps, in what kinds
of proof?  A first suggestion is to look to the rule of Existential
Elimination (\EE) in a Gentzen-Prawitz style natural deduction
system(\cite{Prawitz65}):
%
%
%
%
%

In this style of proof, one does not prove things directly from
existentially quantified sentences $(\exists x)\Phi$; rather one must
assume $\Phi[x/a]$ for some otherwise indeterminate entity labeled
$a$ and then use this assumption, together with whatever other
sentences are available in the proof as so far constructed, to derive
$\Psi$, whereupon one can discharge the assumption.

How is one to think of the labels, typically called parameters,
introduced by (\EE), and eliminated by Universal Introduction (\UI)?
Syntactically they are atomic (individual) terms; but they are not
variables, for they are not bindable, nor are they individual
constants, for they are not to be thought of as in the domain of
interpretation functions.\footnote{Below we shall discuss proof
systems in which there is not a separate set of parameters, but only
individual variables.  In such systems, too, variables introduced by
the rule of Existential Instantiation (\EI) have rather special,
derivation-dependent properties.} Moreover, though this is often not
made explicit, derivations are presumed to involve only {\em
sentences}, formulae with no parameters (and no free variables), as
undischarged assumptions---basic premises---and only sentences as
conclusions.  In a sense, parameters are not expressions in the
language over which the provability predicate or derivation relation
is being defined.  They play a role only in the construction of proofs
or derivations; only {\em in media res}.  Parameters are
proof-dependent.  Indeed, they can actually be shown to be
inference-step dependent, that is, associated with exactly one
application of either \EE or \UI.\footnote{See \cite{Prawitz65},
especially the definition of {\em pure} parameters.}

Some of these features of parameters are analogous to certain
characteristics of discourse markers (especially those associated with
indefinites) in Kamp's theory or indices in Heim's.  In these latter
cases, of course, the dependency involved is to texts or discourses,
not proofs.  Indeed, there is a sense in which these devices lead a
double life, akin to parameters, or to program variables, as we urged
above, {\em during} the processing of a text, but then akin to
quantified variables when the processing of the text has produced the
final output DRS for interpretation.  In these dynamic accounts, then,
attention is drawn to the fact that there are intermediate stages or
steps involved in the production of the final semantically
evaluable/interpretable item (DRS or wff or file).  We shall now turn
to a brief examination of an analogous phenomenon in alternative
derivation systems.

As we noted above, in Gentzen-Prawitz style systems of natural
deduction, the structure of the proof associates with a given
parameter $a$ occurring in the proof a unique inference, an
application of either (\EE) or (\UI), in which that parameter is
involved as the {\em proper parameter} of the particular
application.\footnote{The terminology is Prawitz's.} In fact not all
derivations that meet the required derivation construction
requirements meet this last condition (of {\em purity}), but they can
all be transformed quite straightforwardly into derivations that do.
This is not in general the case for systems of natural deduction that
replace (\EE) with a rule (\EI) of existential instantiation, and
(\UI) with (\UG).

These systems are strikingly different from Gentzen-Prawitz systems in
ways that go beyond the difference in the two troublesome quantifier
rules; we will not go into any detail here.\footnote{The rules of
Existential Introduction (Generalization) and Universal Elimination
(Instantiation) are essentially identical in the two types of system.}
But we do want to make two points about them: (i) derivations in these
systems are, like derivations in Hilbert-style systems, sequences of
formulas, not (proper) trees and (ii) it is much more natural to think
of these `(\EI)' systems as being constructed in one direction, down
from the premises.  Of course, derivations in such systems can be
constructed by a combination of downward (forward) and upward
(backward) directed steps; but one cannot define for them a middle, as
one can for Gentzen-Prawitz derivations in normal form.  Indeed, one
cannot define an interesting normal form for them, either.  They are
not motivated by the kinds of symmetries at the heart of
Gentzen-Prawitz style systems; what Prawitz calls the {\em inversion
principle} and proof reduction (normalization) plays no part in their
design or presentation, or metatheory.

As for the rules themselves, (\EI) is stronger than (\EE); in
particular, one can infer conclusions directly from existentials,
though only under certain restrictions.\footnote{We shall remain
silent on the question of which of these two types of natural
deduction system is a more faithful model of informal deductive
reasoning.} These restrictions tend to be of two sorts, local and
global.\footnote{See Fine's discussion in
\cite{Fine85} of \cite{Quine62}.} The local restrictions have the same
force as the restrictions on (\EE) and (\UI), involving occurrences of
the instantial term (parameter or variable) {\em above} the inference
in question.  The global restrictions, on the other hand, apply only
to a {\em completed} proof or derivation.

For instance, Quine imposes the following {\em ordering} condition on
{\em flagged variables}, variables that play the role of instantial
terms for (\EI) and (\UG) analogous to that of parameters in
Prawitz-style proofs:

\begin{quote}

It must be possible to list the flagged variables of a deduction in
some order $V_{1}, \ldots , V_{n}$ such that, for each number $i$ from
1 to $n-1$, $V_{i}$ is free in no line in which $V_{i+1}, \ldots ,
V_{n}$ is flagged. (\cite{Quine62})

\end{quote}

\noindent This condition (and less troubling {\em flagging condition}) are
required for the proof of soundness to go through:

\begin{quote}
Thus it is that we are unwarranted in supposing the last line of a
deduction to be implied by its premises unless what we have is a {\em
finished} deduction; one whose flagged variables are not free in the
last line nor in premisses of the last line.  (\cite{Quine62}.
Emphasis in the original.)

\end{quote}

There are various ways to interpret this lack of determinacy with
respect to the property of being a sound derivation (lack of
line-by-line soundness).  In Fine's generic interpretation it
corresponds to the fact that one can not determine the meaning of an
instantial term (flagged variable/parameter) until one knows what
application of generalization, especially what application of (\UG),
will eliminate the term and one can't know this with certainty until
al parameters introduced in the derivation have been eliminated by
generalization.  Moreover he notes that if we try to construct the
ordering required by the ordering principle as we proceed with the
construction of the derivation, we may have to {\em revise} this
ordering.

Here we shall take a different perspective, largely because of
connections with work done in the tradition of E-type approaches to
discourse anaphora, especially \cite{Does93}.

\subsection{Epsilon-terms}

Hilbert introduced $\epsilon$-terms and designed the
$\epsilon$-calculus as part of his attempt to justify classical
infinitistic results by purely finitary methods, in particular by
transforming nonfinitistic proofs into finitistic (combinatorial)
proofs (\cite{Hilbert25,Hilbert28}).  Since, on the natural
conception, the existential and universal quantifiers are infinitary
operators (`abbreviations' of infinitary disjunctions or
conjunctions), Hilbert eliminated them in favor of complex singular
terms that can be seen as embodiments of the axiom of choice, which he
took to be central to classical reasoning.  Thus, for each predicate
$A[x]$, he introduced a term $\epsilon x (A[x])$ and an axiom scheme
(the logical $\epsilon$-axiom:

\begin{center}
$A[x] \rightarrow A(\epsilon x (A[x]))$

\end{center}

The scheme justifies the following definitions of the quantifiers:

\begin{enumerate}
\item $(\exists x A[x])^{*} : = A^{*}[\epsilon x A^{*}[x]]$
\item $(\forall x A[x])^{*} : = A^{*}[\epsilon x \neg A^{*}[x]]$
\end{enumerate}

$A$ may contain other free variables, $x_{1}, \ldots$; thus, in
general, $\epsilon x (A[x])$ can be understood as the value of a
choice function, with parameters corresponding to those free
variables.  In what follows, though we shall ignore Hilbert's own
motivation and shall imagine a formalism that includes both
quantifiers and $\epsilon$-terms, and in which the definitions above
are turned into the corresponding inference rules (\EI) and (\UG)---in
the case of (\UG) the definition must be read from right to
left.\footnote{The import of (\UG) is as follows: if even the {\em
ideal} or {\em arbitrarily chosen} non-$A$ item is $A$-ish, then
everything is $A$-ish.}  But this version of these rules requires {\em
no restrictions}, local or global, on occurrences of instantial terms
elsewhere in a derivation.  This system, an abuse of Hilbert's, can be
seen to be a conservative extension over first-order logic,
essentially by way of Hilbert's second $\epsilon$-theorem.

We can now follow a suggestion of \cite{Hazen87} and consider
abbreviating all $\epsilon$-terms in sound derivations by single
letters; the derivations will then look just like derivations in
(\EI)-(\UG) systems.  The restrictions on parameters or flagged
variables then take the form, collectively, of the requirement that
all such instantial terms can be uniquely disabbreviated.  And the
indeterminacy {\em in media res} of the property of being a sound
derivation takes the form of an ambiguity as to what $\epsilon$-term
such a term abbreviates, that is, what the matrix of the term is.

We have noted that the interpretation of a parameter or flagged
variable introduced at a given stage in a (\EI)-(\UG) derivation
depends on what happens subsequently in the derivation. e.g., on what
applications of (\UG) it is involved in.  This is, then, our last
example in the list:

\begin{enumerate}
\item semantic scope of an NP (or of a quantifier)
\item scope of  a program variable
\item extent of an identifier
\item interpretation of a parameter or flagged variable
\item matrix of an $\epsilon$-term

\end{enumerate}

In all these case, we have indeterminacy of the semantic significance
of an expression, relative to (a static analysis of) that part of the
linguistic context $\{$text/discourse/program/derivation$\}$ that has been
encountered up to the point of the occurrence of the expression in
question.  In the case of natural languages and programming languages,
we are dealing with communication tools that are intended to be used
by and for agents of various sorts, with certain known processing
characteristics.  In the case of natural languages, the agents on both
(all) sides of the communicative situation are humans---or machines
programmed to behave, when encountering natural language utterances,
as much like humans in certain respects as is possible.  We often know
how much we can leave indeterminate in a given communicative situation
for we have some pretheoretical sense of the procedures by which we
make the required determinations.  In the case of programming
languages, we can and should fully characterize the various dimensions
of indeterminacy and the procedures by which these are resolved at
run-time.

But what of derivations?  Natural deduction systems, of both of the
kinds we have mentioned, are defined over formal languages, but they
are meant to model certain aspects of informal deductive reasoning.
In particular, the (\EI)-(\UG) systems might be said to be more
faithful models of an aspect of our informal reasoning that we often
ignore or wish away: namely the fact that we often times don't really
know what we're doing or how we should go about deriving a given
sentence from a set of premises.  This is an odd virtue, of course,
and not one that purveyors of such systems are likely to boast about.
Moreover, on the one hand, these systems, e.g., Quine's in {\em
Methods of Logic}, are provably sound and complete and, on the other,
much the same possibilities of false-starts and somewhat blind
proof-search are also available in Gentzen-Prawitz systems.  Still, we
claim that the (\EI)-(\UG) systems, in use in actual proof
construction, more faithfully reflect these features of informal
reasoning than the more elegant and metamathematically satisfactory ND
systems and once again, this is largely due to the feature sketched
above: that the precise derivational import of a line is not in
general determined by the structure of the derivation at the stage
when the line is added.\footnote{Much more can and should be said
about the structure of derivations, about, for instance, the
centrality of the subformula property in constraining derivations; but
it shall have to wait for another occasion.}

\section{Concluding Philosophical Considerations}
\label{sec-conc}

In their discussion, in \cite{GroenendijkStokhof91}, of the Principle
of Compositionality, Groenendijk and Stokhof note that it is primarily
a methodological principle, to be adhered to when possible, to be
eschewed if the empirical or computational or philosophical costs are
too high.  One philosophical consideration has to do with what might
be called the autonomy of semantics, in particular its autonomy with
respect to psychological or cognitive or mentalist issues involved in
theorizing about processes involved in natural language understanding.
``The best way to go about is to carry on semantics as really a
discipline of its own, not to consider it a priori a branch of
cognitive science...''  In particular, they deny the force of the
metaphor that they feel underlies some, at least, of the attraction of
positing a level of mental representation of the content or meaning of
natural language utterances:

\begin{quote}
Our own opinion, for what it is worth, is that the calculating mind is
a metaphor rather than a model.  It is a powerful metaphor, no doubt,
on which many branches of `cognitive' science are based, and sometimes
it can be helpful, even insightful.  But it remains a way of speaking,
rather than a true description of the way we are.
(\cite{GroenendijkStokhof91})

\end{quote}

Quite independent of the computational metaphor for mind, there is
something odd about the claim of the autonomy of the semantics of
natural languages.  Semantics, at least semantics in the
model-theoretic vein of the school of Tarski, is an application of set
theory.  Indeed, the theorems of a Tarski-style definition of truth or
satisfaction are theorems of applied (impure) set theory; they can
even be made into theorems of pure set theory through a little coding.
Set theory is a part of mathematics, and it is certainly autonomous
from psychology (at least it is in my opinion, for what it is worth).
But natural languages and their use in the expression of thought and
in communication are {\em natural}, contingent, empirical phenomena,
involving essentially, or so it would seem, things going on in
people's minds, as well things going on outside their skulls.  This
does not constitute an argument against the autonomy of semantics,
though it might constitute the beginning of one; but it does suggest
that it is not at all clear that one should {\em not} consider the
semantics of natural languages a branch of cognitive science, however
computationally this latter is itself conceived.\footnote{Indeed, it
should be noted that in ``Dynamic Montague Grammar,'' Groenendijk and
Stokhof specify the content of a natural language sentences $\phi$
followed by $\psi$ as ``sets of propositions {\bf p} such that after
the {\em processing} of $\phi$ in state $s$, it holds that {\bf p}
holds after the {\em processing} of $\psi$.''  Surely this is
reference to processing by a mind of some sort.}

Finally, Groenendijk and Stokhof return to the theme of the
inadequacies of natural language that we have seen raised by Frege and
Peano.  Allow me to quote at length:

\begin{quote}
It may be the case, though, that for some the acceptance of level of
logical representation springs forth from a positive philosophical
conviction, viz., a belief in the deficiencies of natural language as
a means to convey meaning.  Now such there may be (or not) when we
consider very specialized kinds of theoretical discourse, such as
mathematics...In such cases, clearly there is room for extension and
revision, for regimentation and confinement.  But that is not what is
at stake here. Here, it turns on the question whether natural language
structures themselves, {\em as we encounter them in spoken or written
language}, then and there are in need of further clarification in
order to convey what they are meant to convey.  In this matter,
semantics, we feel, should start from the premiss that natural
language is all right.  If anything is a perfect means to express
natural language meaning, natural language is.  It can very well take
care of itself and is in no need of (psycho)logical reconstruction and
improvement in this respect.  (\cite{GroenendijkStokhof91})

\end{quote}

How are we to reconcile this view with Frege's?  Assuming that one
important component of the meaning of sentences lies in their logical
interrelationships, and assuming that one can't easily or
systematically determine when one sentence is a logical consequence of
another or of a finite set of others, it does seem that Frege's view
is that natural languages are not perfect means of expressing meaning.
Also, how are to reconcile this view with what Groenendijk and Stokhof
themselves say in \cite{GroenendijkStokhof90}.  Their view in that
paper is that sentences of natural language convey partial information
about the references of discourse markers (of which more later).  But
of course they realize that natural language sentences don't
themselves contain discourse markers, just as they don't themselves
contain indices, variables, or explicit scope indicators.

\begin{quote}
[W]e need to choose a particular discourse marker in the translation
[into the language of DMG], which is indicated by the index that
occurs in the determiner itself.  So the present approach assumes that
we do not translate sentences as such, but indexed structures, i.e,
sentences in which determiners, pronouns, and proper names are marked
with indices.  The function of the indexing mechanism is, of course,
to stake out possible anaphoric relationships among constituents; if a
pronoun is to be related anaphorically to an {\em NP}, a necessary
(but not sufficient) condition is that both carry the same index.

\end{quote}

In fact their earlier approach assumed something similar, though
simpler.  So natural language sentences, just as they are, as we
encounter them in written and spoken language, seem not to be quite
all right---at least not for purposes of systematic semantic analysis.
Yet they are, perforce, all right for us, as users---speakers and
hearers---of our natural languages.  Frege hints at an explanation of
this puzzle: as used, they are produced and meant to be encountered by
intelligent agents---by us---who are either presumed or known to bring
to bear on these encounters a vast store of knowledge, both quite
general and context-specific, and skills of many sorts.  This
presumption of intelligence on the part of one's audience, and the
fact that it is so often shown to be well-grounded, allows speakers to
leave much `to guesswork'; more exactly, the presumption is to the
effect that something much more rational and systematic than mere
guesswork on the part of one's listeners can be depended on.

\section{Conclusion}

Let us return to the original diagnosis of the troublesome phenomena:

\begin{quote}
The phenomena all involve {\em anaphoric pronouns} {\em outside the
scope} of their [indefinite] NP antecedents.
\end{quote}

\noindent This has the advantage of being plausible to linguists, as
it is grounded in widely accepted empirical generalizations; but it
has the disadvantage of being purely negative.  The more general
diagnosis and corresponding positive program of research might be put
as follows:

\begin{quote}
Donkey- or discourse-anaphora shows us that we cannot think of the
semantic values of natural language expressions, in particular of
sentences, as isolated entities, to be generated one per sentence,
though in a systematic way, by a semantic account. Rather our semantic
account must make room for possibilities of semantically significant
interaction among sentences in sequence---for the ways in which the
semantics of a sentence can be effected by sequences of sentences that
are its `prefixes' in a text and the ways in which it can, in turn,
effect the semantics of sentences that come after it.
\end{quote}

This has the ring of illuminating truth, but what truth? It certainly
does not follow that the semantics of sentences cannot be adequately
represented by sets of things, e.g., assignments, rather than sets of
ordered pairs of things.  Perhaps the truth lies deeper: that contra
Frege {\it et al.}, the central unit of semantic analysis is not the
sentence.  Deeper still: the interactions among sentences of which we
speak can only be mediated by, indeed grounded in, the cognitive
activities and processes of agents who use language, by putting
together sentences for certain purposes and inferring those purposes
by understanding the sequences so constructed.

It is time to repeat the main claims of this essay.

Natural languages are used by intelligent agents to communicate with
and influence one another.  The same is true, though in a limited and
unidirectional way, of programming languages.  It is not true, in any
direct way, of the formal languages developed as precise mathematical
tools for analyzing mathematical proof.  It was a crucial requirement
that they wear their meanings on their sleeves---that no guesswork,
inspired or not, be necessary to determine under what conditions a
well-formed sentence belonging to such a language is true.  In our
view, the real import of the phenomenon of donkey anaphora is just
that it is an especially simple and forceful reminder that natural
languages are not like formal languages {\em in that respect}.

The revolution wrought by Montague consists in the fact that he was
the first logician to apply sophisticated techniques of mathematical
logic to a systematic analysis of the semantics of natural languages.
This was especially striking coming from a student of a tradition
according to which natural languages were not fit objects of such
analysis; but in any event the revolution in attitude is more
important than any detailed treatment of particular phenomena.  The
revolution is also more important than, and independent of, the
philosophical attitude of its originator.   Montague, like
Tarski, took semantics to be a branch of applied set theory.  We have
argued that the semantics of natural languages cannot be so taken.

A complete theory of a programming language must include an account,
at some level of abstraction, of the `psychology' of the machines that
execute the language.  Just so, a complete theory of a natural
language must include an account, at some level of abstraction, of the
virtual machines in the heads of its users: {\em us}.\footnote{Given
the sweeping nature of the conclusion, we consign to a footnote the
reminder that in the treatments in \cite{PaginWesterstahl93} and
\cite{Does93} the only aspect of our psychology involved is the fact
that we process texts sequentially, left-to-right.  And, of course, we
are ignoring cataphora and conditionals in which the consequent has
the bad manners to precede the antecedent.  To deal with such cases
adequately requires a story about revisions in the process of
interpretation.}

\section{Acknowledgements}
I would like to thank the participants and audience at the Ninth
Amsterdam Colloquium for much useful comment and criticism on the oral
version (and for not attacking me physically, though the temptation
was great).  Particular thanks in these regards to Jeroen Groenendijk,
Theo Janssen and Martin Stokhof.  Just before visiting in Amsterdam, I
gave a version of the talk at the University of Stockholm, as guest of
Dag Westerst\"{a}hl.  I thank him for putting up with hours of my
ranting and raving about the subjects of this essay and for teaching
me much about them. Thanks also to Peter Pagin, for some of the
same---though he suffered less than Westerst\"{a}hl did.  Finally I am
grateful to my colleagues at SRI for careful attention paid.  Mark
Gawron actually read an early version and the final one owes much to
his perspicacity and tenacity.  Needless to say, they are all
innocent---jointly and individually.


\begin{thebibliography}{99}


\bibitem{GroenendijkStokhof91} J.~Groenendijk and M.~Stokhof,
``Dynamic Predicate Logic,'' {\bf Linguistics and Philosophy}, {\bf
14}, Number 1, 1991, pages 39-100.

\bibitem{GroenendijkStokhof90} J.~Groenendijk and M.~Stokhof,
``Dynamic Montague Grammar,'' in {\bf Proceedings of the Second
Symposium on Logic and Language}, L.~K\'{a}lm\'{a}n and L.~P\'{o}los,
eds., Akad\'{e}miai Kiad\'{o}, Budapest, 1990.

\bibitem{PaginWesterstahl93} P.~Pagin and D.~Westerstahl, ``Predicate
Logic with Flexibly Binding Operators and Natural Language
Semantics,'' {\bf Journal of Logic, Language and Information}, {\bf
2}, Number 1, 1993, pages 89-128.

\bibitem{Kamp81}
H.~Kamp, ``A theory of truth and semantic representation,'' {\bf
Formal Methods in the Study of Language}, J.~Groenendijk, T.~Janssen
and M.~Stokhof, eds., Mathematical Centre, Amsterdam, 1981, pages
277-322.


\bibitem{Montague74a}
R.~Montague, ``English as a Formal Language,'' {\bf Formal
Philosophy}, R.~M.~Thomason, ed., Yale University Press, New Haven,
1974, pages 188-221.

\bibitem{Montague74b}
R.~Montague, ``Universal Grammar,'' {\bf Formal Philosophy}, pages
222-246.

\bibitem{Montague74c}
R.~Montague, ``The Proper Treatment of Quantification in Ordinary
English,'' {\bf Formal Philosophy}, pages 247-270.



\bibitem{vanHeijenoort67}
J.~van Heijenoort, ed., {\bf From Frege to Godel: A Source Book in
Mathematical Logic}, Harvard University Press, Cambridge, 1967.

\bibitem{EijckVries92}
J.~van Eijck and F.~J.~de Vries, ``Dynamic Interpretation and Hoare
Deduction,'' {\bf Journal of Logic, Language and Information}, {\bf
1}, Number 1, 1992, pages 1-44.

\bibitem{Heim83}
I.~Heim, ``File Change Semantics and the Familiarity Theory of
Definites,'' in {\bf Meaning, Use, and Interpretation of Language},
R.~Bauerle, C.~Schwarze and A.~von Stechow, eds. De Gruyter, Berlin,
1983, pages 164-178.

\bibitem{Does93}
J.~van der Does, ``The Dynamics of Sophisticated Laziness,'' Institute
for Logic, Language, and Computation, University of Amsterdam,
manuscript, September, 1993.

\bibitem{Pratt76}
V.~Pratt, ``Semantical Considerations on Floyd-Hoare Logic,'' {\bf
Proceedings of the 17th Ann. IEEE Symp. on Foundations of Computer
Science}, 1976, pages 109-121.

\bibitem{Harel79}
D.~Harel, {\bf First-Order Dynamic Logic}, Lecture Notes in Computer
Science, 68, Springer Verlag, Berlin, 1979.

\bibitem{EijckFrancez93}
J.~van Eijck and N.~Francez, ``Procedural Dynamic Semantics,
Verb-Phrase Ellipsis, and Presupposition,'' Technical Report, 761,
Technion, Haifa, Israel, January, 1993.

\bibitem{Prawitz65}
D.~Prawitz, {\bf Natural Deduction. A proof Theoretical Study},
Almqvist \& Wiksell, Stockholm, 1965.

\bibitem{Fine85}
K.~Fine, ``Natural Deduction and Arbitrary Objects,'' {\bf Journal of
Philosophical Logic}, {\bf 14}, Number 1, 1985, pages 57-107.


\bibitem{Quine62}
W.~V.~O.~Quine, {\bf Methods of Logic}, Holt, Rinehart and Winston,
New York, 1962.

\bibitem{Hilbert25}
D.~Hilbert, ``On the Infinite,'' in \cite{vanHeijenoort67}.

\bibitem{Hilbert28}
D.~Hilbert, ``The Foundations of Arithemtic,''  in \cite{vanHeijenoort67}.


\bibitem{Hazen87}
A.~Hazen, ``Natural Deduction and Hilbert's $\epsilon$-Operator,''
{\bf Journal of Philosophical Logic}, {\bf 16}, Number 4, 1987, pages
411-422.

\bibitem{Hintikka83}
J.~Hintikka (with J.~Kulas), {\bf The Game of Language}, Reidel,
Dordrecht, 1983.


\end{thebibliography}
\end{document}